# A Transformer-based Cross-modal Fusion Model with Adversarial Training for VQA Challenge 2021

Ke-Han Lu*
Department of CSIE
NTUST, Taiwan
M10915010@mail.ntust.edu.tw

Bo-Han Fang*
Department of CSIE
NTUST, Taiwan
B10630012@mail.ntust.edu.tw

Kuan-Yu Chen
Department of CSIE
NTUST, Taiwan
kychen@mail.ntust.edu.tw

## Abstract

*In this paper, inspired by the successes of vision-language pre-trained models and the benefits from training with adversarial attacks, we present a novel Transformer-based cross-modal fusion modeling by incorporating both notions for VQA challenge 2021. Specifically, the proposed model is on top of the architecture of the VinVL model [19], and the adversarial training strategy [4] is applied to make the model robust and generalized. Moreover, two implementation tricks are also used in our system to obtain better results. The experiments demonstrate that the novel framework can achieve 76.72% on the VQAv2 test-std set.*

## 1. Introduction

Visual question answering (VQA) [1, 5], which requires semantically and syntactically fine-grained understanding of images and questions, together with visual reasoning to predict accurate answers, has attracted much attention in recent years. Since a VQA system has to properly integrate different modalities, such as computer vision (CV) and natural language processing (NLP), the stream of research is served as a testbed for building next-generation AI systems.

The bidirectional encoder representations from Transformers (BERT) model [3] has gained much interest due to its state-of-the-art performances in several NLP-related tasks. One of the successes comes from a framework where a pretraining stage is followed by a task-specific fine-tuning stage. Motivated by BERT in NLP, many recent VQA researches [7, 10, 15, 14, 18, 2, 8, 19, 11, 6, 20] start to propose pre-trained models, which are usually trained on large scale image-text pairs [9, 13, 12, 17] in a cross-modality manner with various objectives, and then a VQA system can be deduced by fine-tuning the pre-trained model with some labeled data. Following the research trend, a novel Transformer-based cross-modal fusion modeling, which is based on the architecture of the VinVL model [8, 19] and is trained with the adversarial training method [4], is presented in this paper. Moreover, two implementation tricks, i.e., model average and prediction ensemble, are also introduced to obtain better results.

## 2. Methodology

### 2.1 Transformer-based Cross-modal Fusion Model

**Pre-processing:** Given an image-question pair, a Question-Object-Image triplet $(Q, O, I)$ can be generated, where $Q$ is a sequence of token embeddings $\{q_1, q_2, \ldots, q_{|Q|}\} \in \mathbb{R}^{V \times |Q|}$ of the question, $O$ denotes a series of token embeddings $\{o_1, o_2, \ldots, o_{|O|}\} \in \mathbb{R}^{V \times |O|}$ of object tags that detected from the input image and $I$ represents a set of position-sensitive region representations $\{i_1, i_2, \ldots, i_{|I|}\} \in \mathbb{R}^{V \times |I|}$ extracted from the image, by using the token representations from BERT, a pre-trained object detector, and a simple trainable linear projection. More formally, the object detector can produce 2048 statistics for each RoI, and then the feature is concatenated with a 6-dimensional position encoding of the region. Finally, the 2054-dimensional feature vector is further transformed by a simple linear projection to ensure that it has the same vector dimension $V$ as that of token embeddings.

**Model Architecture:** To answer a question by referring to a given image, a Transformer-based cross-modal fusion (TCF) model is employed to digest text and visual statistics together in this study. Besides, in order to leverage the benefits from large-scale pre-trained models, the TCF model is initialized by the Oscar+ [19]. The input of the TCF model is a set of feature vectors in the form of the original BERT style

$$[[CLS], q_1, \ldots, q_{|Q|}, [SEP], o_1, \ldots, o_{|O|}, [SEP], i_1, \ldots, i_{|I|}],$$

where [CLS] represents a special token of every concatenation feature sequence and [SEP] is a separator token. After going through a series of Transformers, a set of high-level representations $H \in \mathbb{R}^{D \times (3+|Q|+|O|+|I|)}$ is obtained, where $D$ denotes the hidden dimension. Finally, the high-level representation $h_{[CLS]}$ corresponding to the [CLS] token can be treated as a vision-language fusion embedding, so a fully-connected layer with a softmax activation function is stacked on top of it as a classifier to predict the answer for the given question.

---

*The two authors contributed equally to this paper.

## 2.2 Adversarial Training Strategy

To further make the model robust, the adversarial training strategy is employed to refine the proposed TCF model for the VQA task. However, it is generally agreed upon that the adversarial training exemplars are not easy to generate and/or collect, especially for the VQA task. To mitigate the challenge, we turn to perform adversarial training on the embedding level, instead of operating on image pixel and token level as conventional studies [4]. Specifically, in this study, we concentrate on adding adversarial perturbations on the text-modality features only, while the strategy can also be applied to the image-modality. More formally, a set of learnable perturbation vectors $\delta = [\delta_Q, \delta_O] \in \mathbb{R}^{V \times (3+|Q|+|O|)}$ is introduced to disturb each token embedding in the input. Thereby, the first adversarial objective, that aims at making the model robust when the token embeddings have been disturbed, is straightforward

$$\mathcal{R}_{CE}(\theta, \delta) = \text{CrossEntropy}(\text{TCF}_\theta(Q + \delta_Q, O + \delta_O, I), y),$$

where $\theta$ is the set of parameters of the TCF model, and $y$ corresponds to the ground-truth answer. In addition, another objective function is defined as

$$\mathcal{R}_{JSD}(\theta, \delta) = \text{JSD}\left(\text{TCF}_\theta(Q + \delta_Q, O + \delta_O, I), \text{TCF}_\theta(Q, O, I)\right),$$

where JSD denotes the Jensen–Shannon divergence. It is worthwhile to note that the former promotes the model equipped with the robust ability, while the latter advocates that the prediction distribution should also be closed for the original and the perturbed triplets. Consequently, combined with the conventional objective $\mathcal{L}_{con}(\theta) = \text{CrossEntropy}(\text{TCF}_\theta(Q, O, I), y)$, the final training objective can be formulated as a minimax problem

$$\min_\theta \max_\delta \mathcal{L}_{con}(\theta) + \mathcal{R}_{CE}(\theta, \delta) + \alpha \cdot \mathcal{R}_{JSD}(\theta, \delta),$$

where $\alpha$ is a hyperparameter that controls the importance of $\mathcal{R}_{JSD}(\theta, \delta)$. To sum up, the training objective intends to minimize the loss between the ground-truth answer and the model prediction, and it also tries to produce fine-grained competitions for training so as to lead to a more robust model at the same time.

## 3. Experimental Setup & Results

The architecture of the proposed TCF model is identical to BERT$_{base}$ [3] and is initialized by the pre-trained Oscar+$_B$

| Method | Test-dev | Test-std |
|---|---|---|
| VinVL$_B$ [19] | 75.95 | 76.12 |
| TCF | 75.64 | 75.76 |
| TCF + Avg. with 5 models | 75.71 | - |
| TCF + Avg. with 10 models | 75.73 | - |
| TCF + Avg. with 15 models | 75.74 | - |
| TCF + Avg. with 20 models | 75.49 | - |
| TCF + AT | 75.75 | - |
| TCF + AT + Avg. with 5 models | 75.97 | - |
| TCF + AT + Avg. with 5 models | 76.14 | - |
| TCF + AT + Avg. with 5 models | 76.23 | 76.32 |
| Ensemble | **76.65** | **76.72** |

Table 1. Experimental Results on VQAv2 dataset (in Accuracy (%)).

[19]. The object detector of the VinVL model [19] is employed in this study. Based on the pre-trained model parameters, the TCF model is obtained by further fine-tuning the model with the conventional objective $\mathcal{L}_{con}(\theta)$ on the VQAv2 dataset for 20 epochs. After that, the adversarial training criterion is applied for another 20 training epochs. Both the results are listed in Table 1. At first glance, there exists a small performance gap between the proposed TCF model with the VinVL$_B$ [19]. This may be due to the data pre-processing methods and randomness during training. Next, the results also reveal that the adversarial training strategy delivers promising improvements than vanilla TCF, while the performance gains are not as significant as expected. Because of the implementation trick suggested by the end2end ASR toolkit [16], we hence try to average the model parameters from the last few iterations. The results are also summarized in Table 1. Surprisingly, for both settings (i.e., with and without adversarial training), we can obtain better results when we increase the number of models to be averaged. We believe the trick can help to reduce the training bias and make the model more robust. Based on our observations, the adversarial training strategy usually makes the training process much more unstable than only considering the conventional objective, thus the trick of model average can bring more benefits for the former, while only can give little improvements for the latter. To further boost the performance, a majority voting procedure is applied to 3 models and their preliminary combinations. From Table 1, the ensemble strategy can indeed enhance the VQA performance.

## 4. Conclusion

In this paper, we presented a novel Transformer-based cross-modal fusion model with an adversarial training strategy for the VQA task. A series of experiments conducted on the VQAv2 dataset demonstrates that the proposed TCF model can deliver competitive and/or superior results for the VQA challenge 2021.

## 5. Acknowledgments

This work was supported by the Ministry of Science and Technology of Taiwan under Grant MOST 110-2636-E-011-003 (Young Scholar Fellowship Program). We thank National Center for High-performance Computing (NCHC) for providing computational and storage resources.